\documentclass[10pt,twocolumn]{article}

\usepackage{cvpr}
\usepackage{times}
\usepackage{epsfig}
\usepackage{graphicx}
\usepackage{amsmath}
\usepackage{amssymb}
\usepackage{caption}
\usepackage{tabularx}
\usepackage{array}
\usepackage{float}
\usepackage{color,soul}

\usepackage[pagebackref=true,breaklinks=true,colorlinks,bookmarks=false]{hyperref}
\usepackage{amssymb,amsmath,amsthm,enumitem}

\cvprfinalcopy 

\def\cvprPaperID{1141} 

\ifcvprfinal\fi
\begin{document}

\title{Disentangled Image Generation Through Structured Noise Injection}

\author{Yazeed Alharbi\\
King Abdullah University for Science and Technology (KAUST)\\
{\tt\small yazeed.alharbi@kaust.edu.sa}
\and
Peter Wonka\\
KAUST\\
}

\twocolumn[{%
\renewcommand\twocolumn[1][]{#1}%
\maketitle
\begin{center}
    \centering
 \includegraphics[width=1\linewidth]{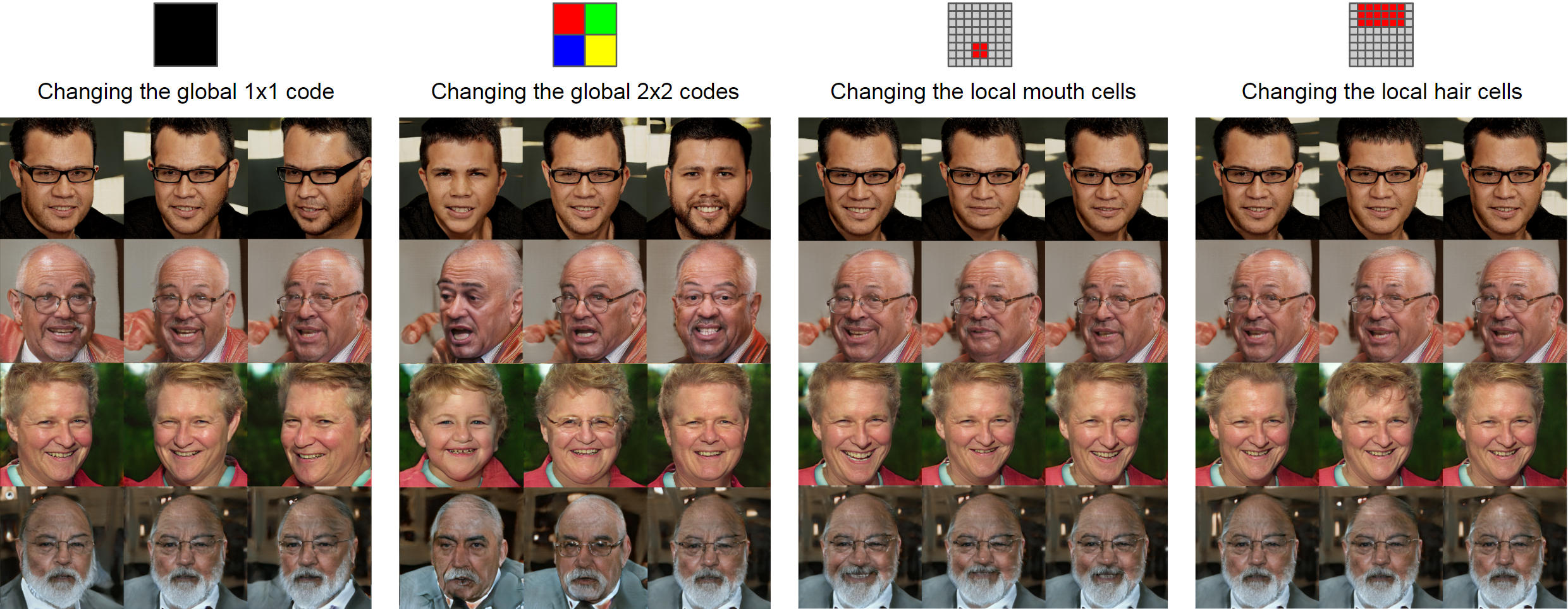}
\captionof{figure}{An illustration of our structured noise injection for image generation. Our method enables changing global and local features in a disentangled way with minimal overhead and without requiring labels.}
\end{center}%
}]

\setlength{\abovecaptionskip}{-0pt}
\setlength{\belowcaptionskip}{-10pt}

\begin{abstract}
\vspace*{-.5em} 

    We explore different design choices for injecting noise into generative adversarial networks (GANs) with the goal of disentangling the latent space. Instead of traditional approaches, we propose feeding multiple noise codes through separate fully-connected layers respectively. The aim is restricting the influence of each noise code to specific parts of the generated image. We show that disentanglement in the first layer of the generator network leads to disentanglement in the generated image.
    Through a grid-based structure, we achieve several aspects of disentanglement without complicating the network architecture and without requiring labels. We achieve spatial disentanglement, scale-space disentanglement, and disentanglement of the foreground object from the background style allowing fine-grained control over the generated images. Examples include changing facial expressions in face images, changing beak length in bird images, and changing car dimensions in car images. This empirically leads to better disentanglement scores than state-of-the-art methods on the FFHQ dataset.
\end{abstract}
\section{Introduction}

    Recent advances in generative modeling using GANs lead to incredible results for synthetic image generation. Most notably, StyleGAN \cite{StyleGAN} is able to generate very high quality images that can be hard for untrained humans to identify as fake. \par
    Improving disentanglement is an open area of research as one of the main criticisms of state-of-the-art GANs is the difficulty of controlling generated images. The goal is to change certain attributes of the generated image without changing the other attributes. For example, it would be desirable to be able to add smile to a face image without changing the identity or the background. \newline
    Current methods for disentanglement are either too limited or too specific. Disentanglement in StyleGAN \cite{StyleGAN} is mainly scale-based. Low-level features can be changed while maintaining high-level features, but it is incredibly difficult to change specific attributes individually. On the other hand, HoloGAN \cite{HoloGAN} disentangles pose from identity, but it uses a specific geometry-based architecture that does not apply to other attributes. \par
    Current GAN architectures map a decorrelated input noise code to an intermediate representation that defines the generated image. There are two main approaches to generating an image from the intermediate representation. The first approach maps the code using fully-connected layers to obtain a tensor with spatial dimensions that is upsampled and convolved to generate an image. The second approach starts with a common input tensor, and uses the input code to modulate the feature maps in a spatially-invariable manner. Both approaches are inherently entangled structures, i.e. every element of the latent code can influence every part in the generated image. \newline
    We argue that a general yet fine-grained form of disentanglement can be achieved through better structuring of noise code injection. Specifically, we design our network such that each part of the input noise controls a specific part of the generated image. First, we propose separating spatially-invariable concepts from spatially-variable ones. To achieve this, we utilize two input codes: A spatially-invariable code and a spatially-variable code. The spatially-invariable code is used to compute AdaIN \cite{AdaIN} parameters. It operates the same way on each pixel within a feature map regardless of location. The spatially-variable code produces the input tensor to the upsampling and convolution layers of the generator. It shows high spatial correspondence with the final image. Second, we propose using a structured spatially-variable code to enable controlling specific regions of the generated images. The spatially-variable code contains codes that are specific to each location, codes that are shared between some locations, and codes that are shared between all locations.
    The benefits of our contributions can be summarized as follows:
    \begin{enumerate}
        \item Greater degree of control over the generated images without requiring labels
        \item Spatial disentanglement of the latent space through grid-based structures
        \item Scale-based disentanglement of the latent space through shared values 
        \item Foreground disentanglement through separating style from spatially-variable values
    \end{enumerate}

        \begin{figure*}
        \centering
        \includegraphics[width=1\textwidth]{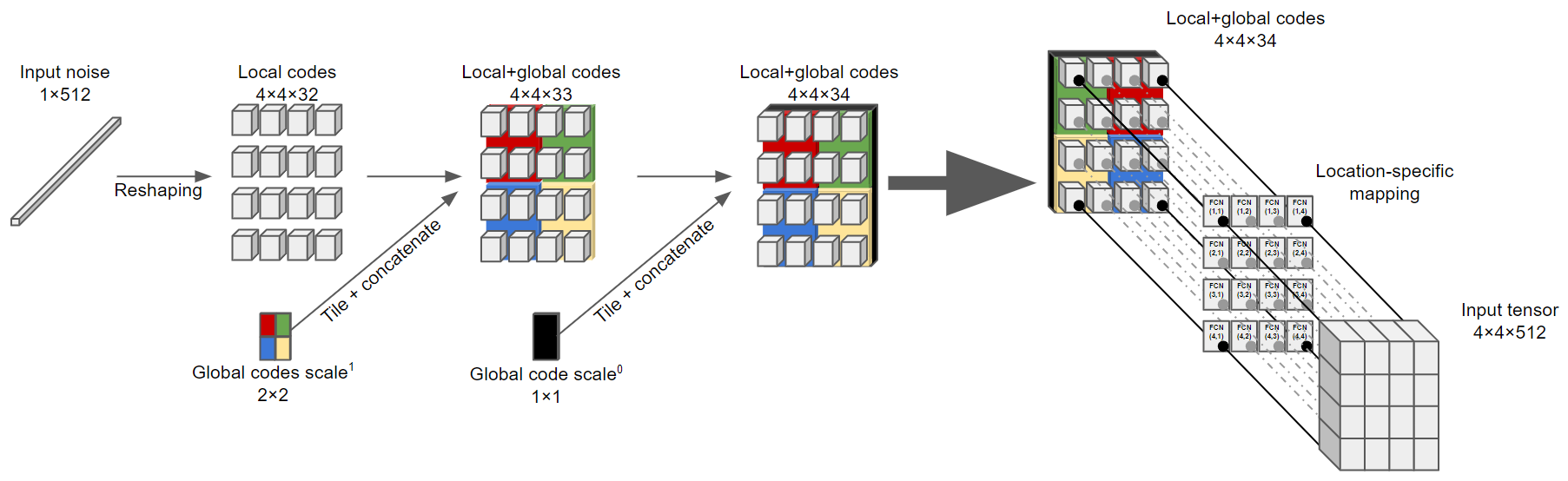}
        \caption{An illustration of the spatially-variable code mapping in our method. Our noise injection structure utilizes separating mapping parameters per code grid cell. Each cell contains a mixture of unique location-specific codes, codes that are shared with neighbors, and codes that are shared with all cells. We show that disentanglement in the input tensor leads to disentanglement in the generated images.}

        \label{localglobalDemo}
         \end{figure*}
\section{Related work}
    \textbf{Methods of noise injection: } The original approach of utilizing the input noise is still a popular option~\cite{GAN, DCGAN, BIGGAN, ProgressiveGAN}. The input noise, which is a vector with many entries, is mapped using a single linear layer to a tensor with spatial dimensions which is then upsampled and convolved to obtain the final image.\newline
    Recent approaches \cite{StyleGAN, FilterScaling, MUNIT} diverge from the original design in two ways: they propose using a deeper fully-connected network to map the input noise vector $z$ into a latent code $w$, and they utilize $w$ to modulate the feature maps by scaling and addition.\par
    \textbf{Face image manipulation: } The majority of face manipulation methods work in image-to-image settings and require semantic labels. Hence, those methods \cite{StarGAN, ResidualFaceManipulation, MaskGAN, IdentityAwareTransfer, ELEGANT, AttGAN} almost universally train on the CelebA datasets because it contains labeled attributes. For example, MaskGAN \cite{MaskGAN} requires labeled masks with precise hand-annotation, and ELEGANT \cite{ELEGANT} uses the attribute labels as an integral part of the network's architecture. Other similar methods \cite{GANimation, TPGAN} use specific datasets with labels suitable for the specific task. We show that this is an unnecessary and undesirable restriction for manipulation of unconditionally-generated images. Especially with the availability of more recent unlabeled datasets that exhibit more variety, such as the FFHQ dataset \cite{StyleGAN}. 
    The common limitations of current face image manipulation methods are the necessity of labels, the necessity of complex adhoc architectures, the degradation in quality, and the inability of restricting changes to local area.\par
    \textbf{General disentanglement: } For general disentanglement, there are two common approaches: background-foreground disentanglement, and entanglement of semantic concepts.
    One of the earlier methods for foreground-background disentanglement \cite{CompositeGAN} incorporates the learning of alpha masks into the network. They propose using several generators to generate RGBA images to blend. FineGAN \cite{FineGAN} shows impressive results on disentangling generation into three stages: background, foreground, and foreground detail. The main drawbacks for previous foreground-background disentanglement methods are the lower quality and variety of generated images, the complexity of the proposed networks, the inability to generalize to disentangle more regions, and the requirement of labels. \newline
    Another approach towards disentanglement is to allow changing specific concepts independently, without enforcing that they should only affect certain pixels of the generated image. The majority of methods following this approach focus mainly on pose \cite{HoloGAN, PersonImageGAN, PoseGuidedGeneration, IdentityPoseGAN, PoseAppearanceGAN}. The disentanglement in this case is often application-specific. Methods that offer a more general disentanglement \cite{LatentController} often suffer from a lower quality of images.\par 
    
    We extend on previous noise injection methods by proposing a structured noise injection method that leads to a more disentangled representation. We extend on face image manipulation methods by not requiring labeled attributes, simplifying the network architecture, allowing user-defined granularity of control, and maintaining the state-of-the-art quality. We extend on previous disentanglement methods by enabling changing spatially-local aspects, changing global aspects, and changing background style independently.\newline
    While face image manipulation is often considered as a separate problem from general disentanglement, our method is general enough to apply to many different datasets with similar results. We use our method to train on car and bird images. We show that even for unaligned datasets containing different objects, our method still offers a high degree of control over the generated images.

\section{Structured noise injection}
\subsection{Motivation and intuition}
\begin{figure}[H]
        \centering
        \includegraphics[width=0.48\textwidth]{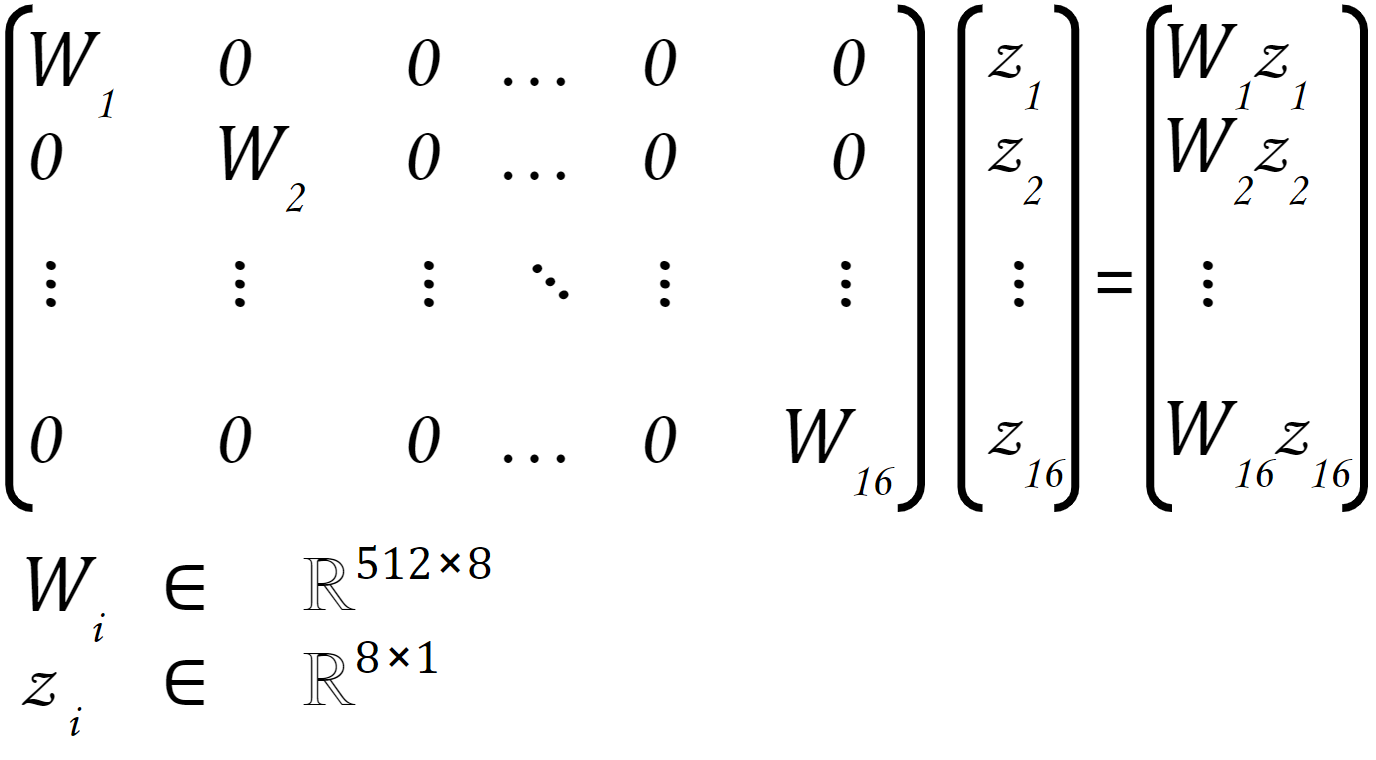}
        \caption{We map each part of the noise code using independent parameters. The result of the mapping is then reshaped to a $4\times 4\times 512$ tensor where each pixel is generated from an independent code that was transformed using independent parameters.}

        \label{fig:ourMapping}
         \end{figure}

    There are several motivating observations for our method.
    The first observation is the evident difficulty of controlling the output image of state-of-the-art GANs. Previous methods map the input noise through a linear layer or several fully-connected layers to produce a tensor with spatial dimensions. We refer to this tensor as the input tensor, as it is typically the first input to the upsampling and convolution blocks of GANs.
        \begin{figure}[H]
        \centering
        \begin{equation}
          InputTensor = Wz + b
        \label{tradEq}
        \end{equation}
        \begin{tabular}{@{}>{$}l<{$}l@{}}
            W \in \mathbb{R}^{(4\cdot4\cdot512) \times 128} \\
            z \in \mathbb{R}^{128 \times 1} \\
            b \in \mathbb{R}^{512 \times 1}\\
            InputTensor \in \mathbb{R}^{4\times 4\times 512}
        \end{tabular}

        \end{figure}
    As shown in equation \ref{tradEq}, traditional methods learn a matrix $W$ to map the entire input noise to vector which is then reshaped to have width, height, and channels. We believe that this design choice is inherently entangled since each entry of the input noise is allowed to modify all spatial locations in the input tensor. This observation led us to explore utilizing separate noise codes per spatial location, essentially limiting communication between spatial sources of variation. We find that by simply dividing the input tensor into regions, providing a different noise code per region, and a separate mapping from each code to the corresponding region, we are able to obtain spatial disentanglement. We restructure the mapping in our method such that $z$ consists of independently sampled parts that are each mapped using an independent part of $W$. The $W$ in this case is sparse as shown in figure \ref{fig:ourMapping}.

    There is a connection between what we observed and the per-pixel noise in the original StyleGAN paper where the authors allude to the effect of spatially-varying values. The authors argue that the network needs spatially-varying values in order to generate stochastic variation. They demonstrate that adding per-pixel random noise allows the network to generate exactly the same face but with minor variation in hair bangs and skin. However, those changes only affect minor details in StyleGAN. We expect that this is due to inserting noise in each level where the network cannot do long-term learning using the noise values. A key contribution of this paper is showing how the noise injection structure can lead to spatial disentanglement.\par
    The second observation is related to style transfer methods and it further guides our specific choice of noise injection structure.
    The style transfer problem is often divided into two competing components: preserving original content and adding exemplary style. Content corresponds to the identity of the image (a specific kind of car, or a building with a specific layout), and it usually affects the loss through per-pixel distances from the generated image features to the input content image features. Style, on the other hand, corresponds to concepts such as color scheme and edges strokes. Style affects the loss through some measure of correlation between generated and exemplary feature map statistics often through the computation of gram matrices. It is easy to see that the content loss is much more dependent on the arrangement of spatial values than style. This is because content loss is based on location and it is often measured using an L2 distance while style loss is based on summary statistics of entire feature maps and it is often measured using correlation. 
    Another key contribution of this paper is proposing a noise injection structure that leads to disentanglement between the foreground object (the spatially-variable content) and the background type (the feature-map-wide style), such that changing spatial regions in the foreground does not affect the background style.\par

       \begin{figure*}
        \centering
        \includegraphics[width=1\textwidth]{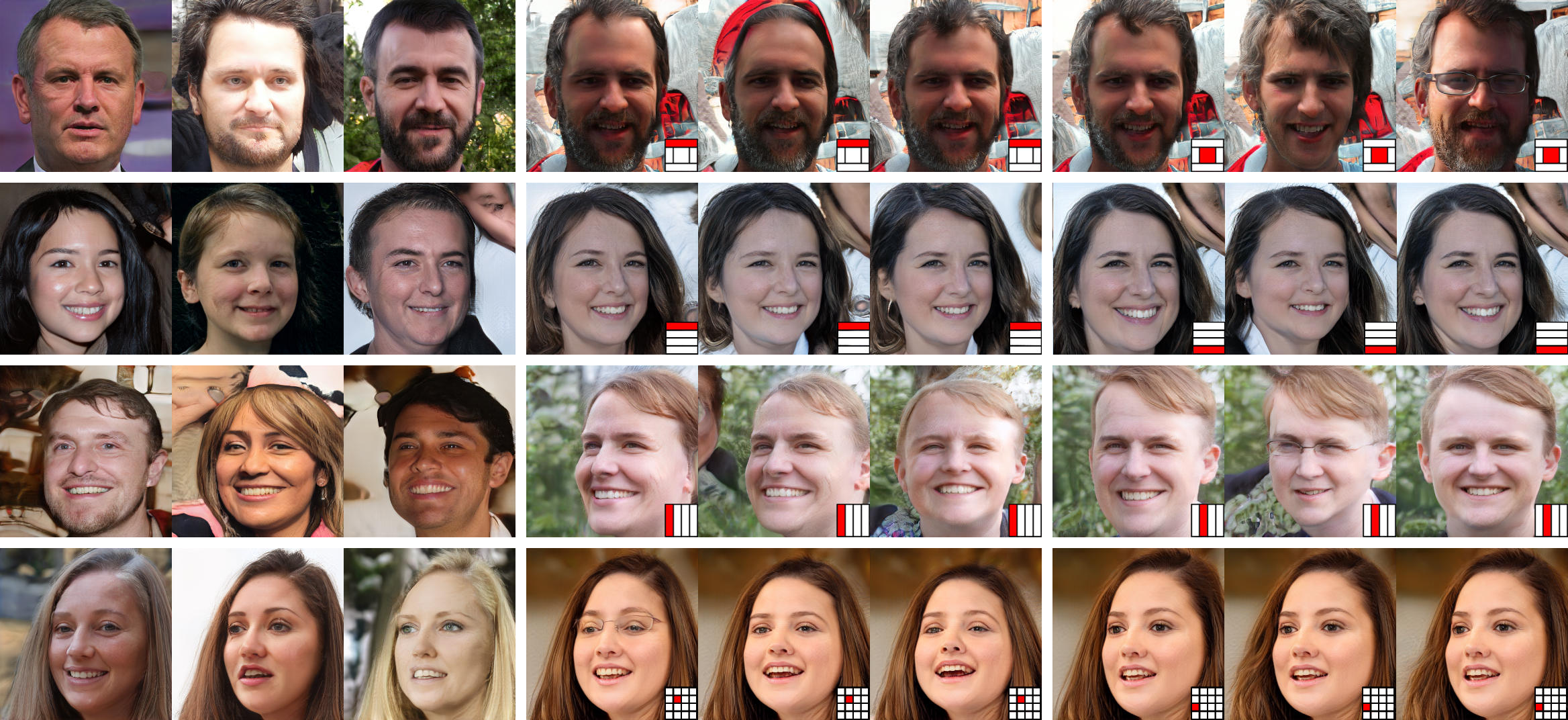}
        \caption{A comparison of different noise injection structures. Rows showcase different injection structures, and columns showcase the results of changing specific parts of the input codes. In the first row we inject a noise code for each region of a manually designed structure. The second and third rows inject a noise code per row and per column respectively. In the fourth row we inject a noise code per input tensor pixel. The first set of columns show the result of changing the style code while keeping local codes fixed. The second and third set of columns show the result of changing some local codes while keeping other local codes and the style code fixed.}

        \label{fig:structures}
         \end{figure*}
         
    \subsection{Structure design space}
    Numerous factors must be considered when designing the noise injection structure.
    One factor is the complexity of the mapping. Some methods \cite{DCGAN, BIGGAN} use a single linear layer to move from noise to latent space, while other methods \cite{StyleGAN, FilterScaling} employ a fully-connected network with several layers.
    Another factor is the way the noise code is used, whether it is used to generate an input tensor to upsample and convolve or used to modulate feature map statistics. 
    Finally, since we propose supplying a noise code per region, a decision has to be made regarding the specific subdivision of the input tensor into regions.
    We find that sampling independent noise codes spatially and pushing them through independent mapping layers is sufficient to achieve disentanglement. We compare several noise injection structures visually and numerically. In Figure \ref{fig:structures}, we show how the particular choice of noise injection structure affects user control over the generated images. \par
    We opt for injecting two types of noise codes combining the benefits of traditional and recent interpretations of noise codes. One code is a style code that is used to modulate feature map statistics. The other code is a spatially-variable code that is used in the traditional way to generate a tensor with spatial dimensions to feed to the upsampling and convolution blocks. The spatially-variable code contains an independent local code per pixel of the input tensor, as well as shared codes. Our approach injects an independent noise code per each pixel (cell) of the input tensor, $2\times2$ codes that are shared per sub-region, and one code that is shared for all cells. Each cell code is pushed through a single linear layer to produce the corresponding pixel in the input tensor. We compared starting from a $4\times4$ and an $8\times8$ input tensor, and find that the latter offers more control. \par 
    The best structure we propose offers local and global disentanglement as well as disentanglement of the foreground object from the background style. We find that it performs equally well regardless of the dataset and whether it is aligned. Across different kinds of datasets, we obtain a significant degree of control ranging from resizing a bird's beak to changing the viewpoint of a car image.

\subsection{Foreground-background disentanglement}
    The first aspect of disentanglement obtained by our method is between the background style and the foreground of an image. This disentanglement is a consequence of our structure of separating spatially-variable sources of variation from feature-map-wide sources of variation. By providing independent spatially-varying values, allowing them to completely define earlier layers, and applying style modulation in later layers, we encourage the spatially-variable code to define the foreground, and the style code to define background and general image appearance. Our formulation has the same effects observed by previous researchers in style transfer; applying the style earlier in the network diminishes the importance of content, and allows the style to be more influential. On the other hand, applying style earlier allows style to change not only the background, but many aspects of the foreground as well. Interestingly, we find that style will always contain most information about the background regardless of whether it is applied in earlier or later layers.
    
\subsection{Local disentanglement}
    The second aspect of disentanglement is between local areas in the foreground. One of the main reasons why our method is able to disentangle concepts is due to our noise-to-tensor mapping. We view the mapped input tensor as a grid where each cell is the result of pushing a location-specific part of the input noise through a location-specific mapping function. As a result, our method enables changing a single cell to change the corresponding part of the generated image. While these changes are localized with high fidelity, it is expected that the changing of one cell can result in minor modifications of surrounding regions due to the spatial overlap in the generator layers.\par

\subsection{Local-global disentanglement}
    The third aspect of disentanglement is between local and global effects. There are drawbacks to using only local values to determine the final image. Since some concepts in the image are global by default (such as pose and gender), we find that the network is forced to associate global concepts with many local values. For example, we noticed that changing a grid cell containing part of the mouth or the jaw can sometimes change pose and gender too which is undesirable. It would be more user-friendly to segregate the global aspects of the image from local ones, such that changing cells containing the mouth may change smile or facial hair but not pose or age.\par
    We propose reserving certain entries of the spatially-variable noise code to be global and shared across local cells. So before the noise-to-tensor mapping, we concatenate the global code with each local code. This is not too different from FineGAN's approach of concatenating the parent code with the detail layer's code. It is intriguing to find that, without any supervision, the network learns to associate the global entry with pose when training on FFHQ face images. We performed several experiments with varying lengths of the global code, and the main caveat is that more expressiveness in the global code leads to lower influence of the local codes and more entanglement of the global code. We opt for allowing only a single value to be shared across all pixels.\par
    After using a single shared entry, and disentangling pose from local features, some issues still remain. Changing the mouth cannot change the pose anymore, however it is still free to change the size of the jaw making the person seem younger or older. We loosely draw inspiration from scale-space theory to add another scale of shared global noise code entries. In order to restrict mid-level aspects in addition to pose, we employ a $2\times2$ code shared with local codes based on location. As shown in figure \ref{localglobalDemo}, local cells at the top left quarter of the input tensor will share the first entry, local cells at the top right quarter will share the second entry, and so on. At that scale, only coarse information about the geometry of a face can be encoded. We experimented with adding another scale level of $4\times4$ but found that it reduces the variance of changes performed at the $8\times8$ scale.

\section{Architecture and implementation details}
    
\subsection{Overview}
    Although our method is in many aspects orthogonal to state-of-the-art networks, there are two design choices of our method that affect the architecture. First, we utilize two input noise codes: a spatially-variable code containing a mixture of local codes and shared codes, and a style code that is used to compute AdaIN \cite{AdaIN} parameters. Second, we utilize a noise injection structure for the spatially-variable code that leads to disentanglement in the input tensor and the generated images.
\subsection{Input noise codes}
    The spatially-variable code in our network is a concatenation of the global code entry, the $2\times2$ entries that are shared by region, and the $8\times8$ local codes for each cell. After structuring and mapping this code to a tensor using our method, the resulting tensor is upsampled and convolved to generate the final image. Another noise code is used to apply style modulation to feature maps. This code is used to learn AdaIN parameters, similar to StyleGAN \cite{StyleGAN}. As mentioned earlier, the utilization of two different input noise codes that operate on different aspects of generation is essential to disentangle the foreground object from the background style and general appearance. \newline 
    For our top-performing structure, the global code dimension is set to 1, the $2\times 2$ region-based codes are each of dimension 1, and the local cell codes are each of dimension 16. So at each cell, in addition to the 16 local code entries, the cell also received the global code entry and the region-based shared entry. Style modulation is applied starting at layer 5 (the second block at resolution $16\times16$), such that the spatially-variable code completely determines the earlier layers.
    

\section{Results}
\subsection{Comparative experiments}
    We perform two sets of experiments: qualitative and quantitative. We use the FFHQ dataset \cite{StyleGAN}, the LSUN cars dataset \cite{LSUN}, and the CUB dataset \cite{CUBDataset}.\par
    For our qualitative experiments, we train on datasets of different objects. We do not utilize any labels or alpha masks. We use our qualitative results to demonstrate that we have achieved our goal of fine-grained category-independent control over the generation process. For face images, we show how our method can change smile, hairstyle, pose, age, and accessories. Although a grid-based system seems more appropriate for aligned datasets such as the FFHQ dataset, the same effects are obtained for unaligned datasets. For car images, we show how we can change camera viewpoint, dimensions of the car, and local details. For bird images, we show how we obtain control over the general colors, the background, and even beak length.  \par

    \begin{table}
    \begin{center}
    \begin{tabularx}{0.48\textwidth}{ | l | l | X | X |} 
    \hline
        Method & FID & $Z$ linear separability & $W$ linear separability \\
    \hline 
       StyleGAN 8 W  & 4.40 & 165.26 & 3.79\\ 
       StyleGAN 2 W  & 4.43 & / & 6.25\\ 
       ProgressiveGAN  & 5.25 & / & 10.78\\ 
    \hline 
    \hline 

        Ours (w/o per-pixel noise)  & 5.92 & 45.36 & 4.29 \\

    \hline
    \end{tabularx}
    \end{center}
    \caption{A comparison of quality and linear separability scores on FFHQ at full resolution. Lower FID indicates higher quality, while lower linear separability indicates more disentanglement.}
    \label{tab:FinalResults}
    \end{table}
   
    For our quantitative experiments, we use the FID for measuring quality, and path length and linear separability for measuring disentanglement. These two disentanglement metrics were proposed by the StyleGAN paper \cite{StyleGAN}. Path length uses the smoothness of successive images generated from interpolated codes as an indicator for disentanglement. Methods that introduce new concepts not present in the two endpoints of the interpolation will have high path lengths and low disentanglement.
    Linear separability works after some processing of generated images. After face images are generated by a method, they are pushed through a pre-trained attribute classification network. Then, the most confident predictions are used as a training set for a linear SVM classifier. The idea is that if the latent codes are disentangled then we can linearly separate the attributes of resulting images using only the codes.\par 
    In Table \ref{tab:FinalResults}, we compare our results with StyleGAN after training on the $1024\times 1024$ FFHQ images.
        \begin{figure}[H]
        \centering
        \includegraphics[width=0.3\textwidth]{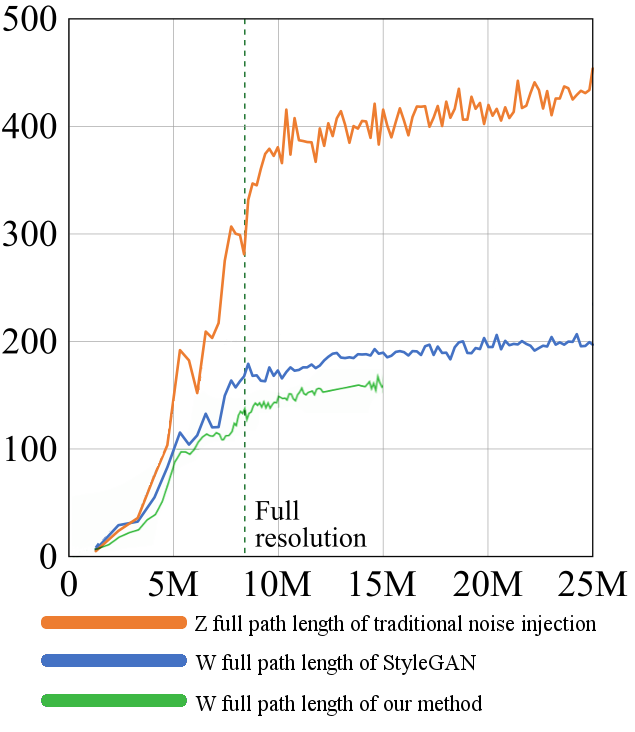}
        \caption{A plot of the W full path length as training progresses. Throughout training, our method leads to more disentanglement as indicated by the shorter path lengths. We adapt the other results from the StyleGAN paper \cite{StyleGAN}}

        \label{fig:pathlengthovertime}
         \end{figure}

\subsection{Ablation studies}
    We perform several experiments to highlight the effects of different components on the performance of our method. We test the effect of adding per-pixel noise, the effect of applying the style code starting at different resolutions, and the effect of the dimension of the local codes. Since we performed a large number of tests with different settings, we opted to compare them on lower resolution $256\times256$ FFHQ images after training until a total of 10 million images is seen. \newline
    In practice, as long as different noise code entries are mapped using different functions for each pixel in the input tensor, then disentanglement will be achieved. We also find that whether style modulation is applied in all layers or only in the last few layers, our method still achieves some foreground-background disentanglement. \newline
    However, the foreground-background disentanglement is dependent on where we begin style modulation. Applying the style modulation too late in the network would force the spatially-variable code to learn to change the background, while the style code will be restricted to mainly changing colors. We believe that this is a useful effect since it could be used to adapt the identity-style decomposition for a variety of image generation problems. \par 
    In table \ref{tab:ArchAblation}, we quantify some of the effects of each component of our method. 
 
        \begin{figure*}
        \centering
        \includegraphics[width=1\textwidth]{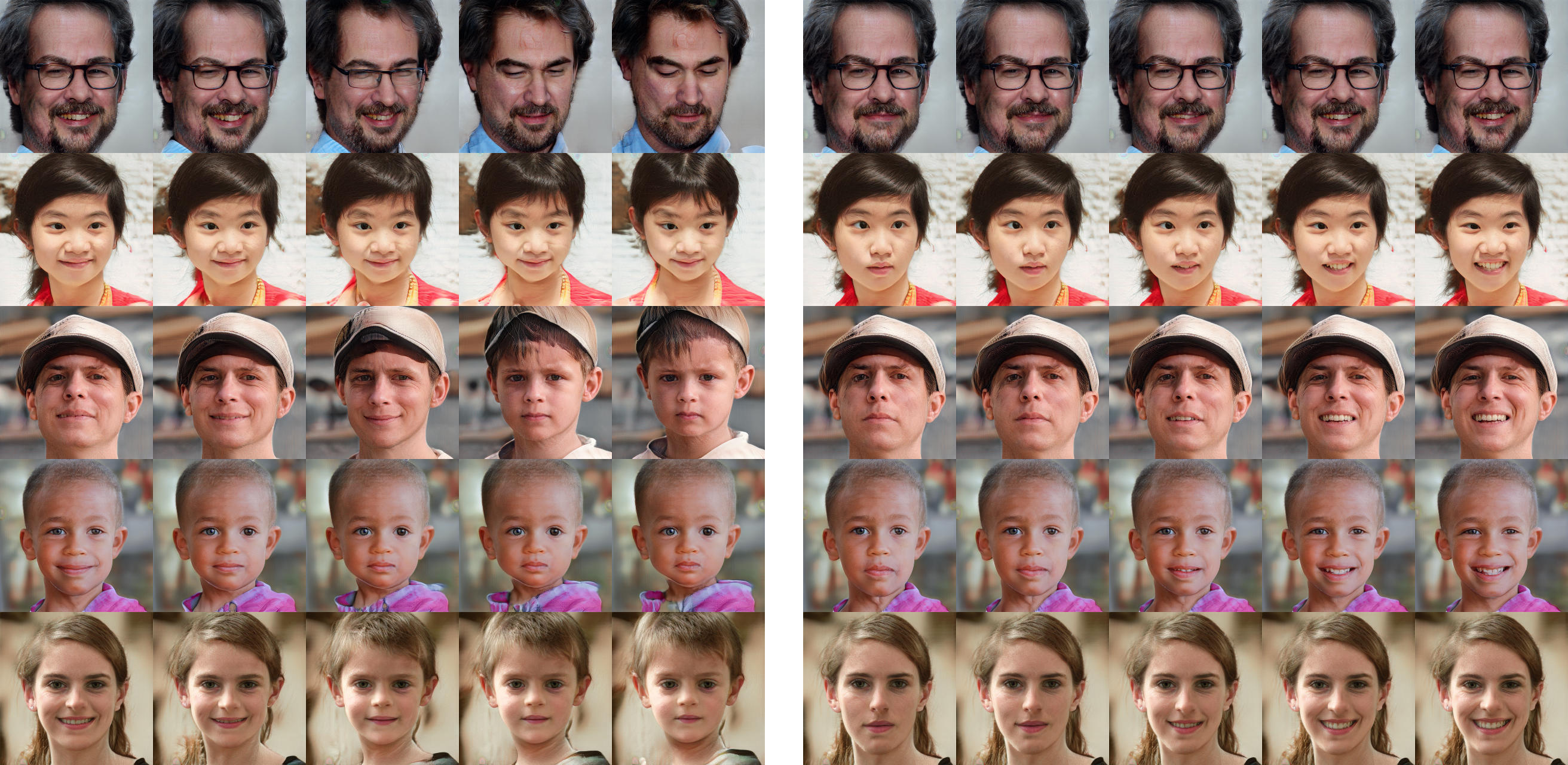}
        \caption{Our network learns attributes in a disentangled way. Left: we sample two scale\textsuperscript{1} global codes. Right: we sample two codes for the four cells surrounding the mouth. For each figure, we use the same two codes to replace and interpolate only the corresponding entries in the full row code. We find that regardless of how each row code is unique, replacing the mouth cells with the same values leads to the same degree of smile.} 
        

        \label{fig:AxesOfVariation}
         \end{figure*}

        \begin{figure}
        \centering
        \includegraphics[width=0.48\textwidth]{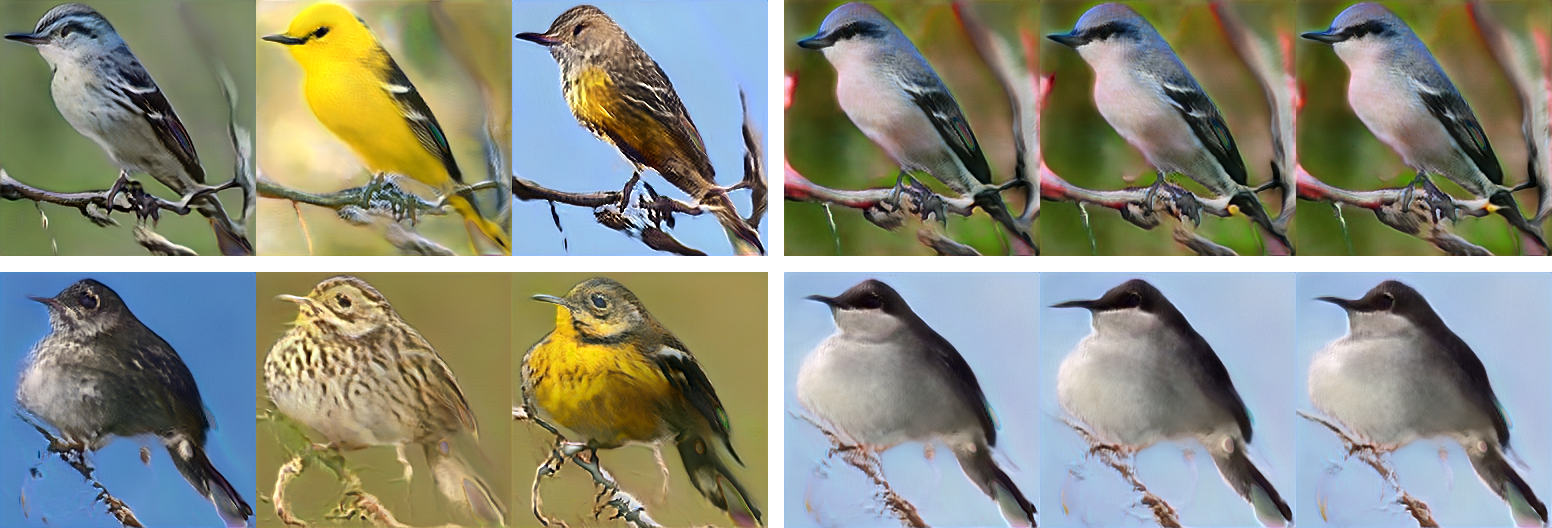}
        \caption{An example of the disentanglement of our method on the birds dataset. We can change background style and color without changing content (left) and change the beak without changing the background style (right).}

        \label{fig:birdImages}
         \end{figure}
         
        \begin{figure}
        \centering
        \includegraphics[width=0.48\textwidth]{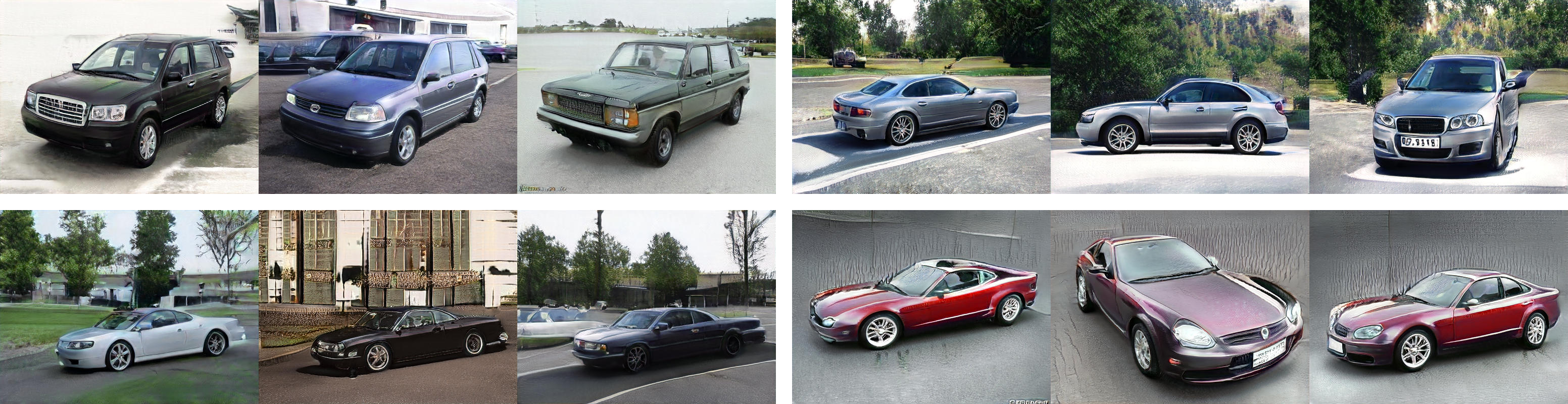}
        \caption{An example of the disentanglement of our method on the cars dataset. We can change background style and color without changing viewpoint (left) and change the camera viewpoint without changing the background style (right).}

        \label{fig:carImages}
         \end{figure}

    \begin{table}
    \begin{center}
    \begin{tabularx}{0.48\textwidth}{ | l | X | X | X |} 
    \hline
        Method & FID & $Z$ path length & $W$ path length\\
    \hline 



        
       StyleGAN w/o per-pixel noise & 10.40 & 546.06 & 176.85\\ 
    \hline 
        Ours w/o per-pixel noise & & & \\
        Style at all layers & 11.55 & 258.69 & \textbf{106.56} \\ 
        Style starting at $16\times16$ & 11.60 & \textbf{208.04} & 111.13 \\ 
        Style starting at $64\times64$ & 13.42 & 215.54 & 158.26\\ 
        Style starting at $128\times128$ & 19.80 & 628.73 & 577.33\\ 
    \hline
    \hline 
       StyleGAN   & 10.73 & 280.58 & 89.91\\ 
    \hline 
        Ours  & & & \\
        Style at all layers & 15.98 & 128.19 & 52.26 \\ 
        Style starting at $16\times16$ & 12.08 & \textbf{93.05} & \textbf{49.14} \\ 
        Style starting at $64\times64$ & 16.05 & 302.39 & 274.59\\ 
        Style starting at $128\times128$ & 16.36 & 276.79 & 239.90\\ 

    \hline
        Ours   & & & \\
        Local code dimension 8 & 12.61 & 103.82 & 55.67 \\ 
        Local code dimension 16 & 12.08 & 93.05 & 49.14 \\ 
        Local code dimension 32 & 11.44 & 108.44 & 51.62\\ 
        Local code dimension 128 & 11.41 & 125.43 & 52.77\\ 
    \hline
    \end{tabularx}
    \end{center}
    \caption{A comparison of different architectures of our method. A smaller path length indicates more disentanglement.}
    \label{tab:ArchAblation}
    \end{table}
\subsection{Analysis}
    To the best of our knowledge, there are no previously established qualitative results for spatial control over the generated images. For the CUB dataset, previous disentanglement results were established by FineGAN \cite{FineGAN}. However, it is limited in two ways: it requires background labels, and it disentangles generation discretely into three categories: background, foreground, and foreground detail. While our method is not able to change the background without changing color of the foreground (since we do not use alpha mask labels), we present the user with a much higher degree of control. On the birds dataset, our method learns without supervision to assign the style code to the color of the bird and the type of background, the global codes to the pose and shape of the bird, and the local codes to local features such as the length of the beak.\par
    We demonstrate an important advantage of our method in Figure \ref{fig:AxesOfVariation}. First, we generate a unique face image per row by sampling a unique style code and full spatially-variable code (1 + $2\times 2$ + $8\times8\times16$) for each row of the figure. Then, for the right figure, we resample two local codes for the cells covering the mouth (2 * ($2\times2\times16$)). The same two codes are used to replace each spatially-variable code at the mouth cells (right). For the left figure, we sample two the scale\textsuperscript{1}  (2 * ($2\times2$)). Again, the same two codes are used to replace the scale\textsuperscript{1} codes of each spatially-variable code (left). We interpolate between the two sets of codes, and show the generated images. It is interesting to note that the same smile code applies the same degree of smile without any dependence on the unique spatially-variable code of each row in the figures. Similarly, we observe that the same intermediate global codes lead to a predictable aging effect. This suggests that our network learns concepts in an abstract way. \par
    In almost all settings and resolutions, our method leads to a substantial improvement in disentanglement scores when compared with StyleGAN. The $Z$ space is highly relevant since our method allows resampling in $Z$, while the $W$ space is used only for interpolation and cannot be sampled easily. Although our method results in a slight increase in FID, we believe that this is an unavoidable cost for disentanglement. Our method in essence enforces additional constraints over traditional GANs. By limiting communication between local cells, the consistency between adjacent cells has to be enforced later in the convolutional blocks of the network. On the other hand, traditional method allow the first mapping function to see the whole noise code and produce a tensor where all the cells see the same input.

\section{Discussion and limitations}
    In this paper, we explore improving disentanglement through better noise injecting structures. In doing so, we contribute to theory and application.\newline
    In terms of theory, our analysis shows that the mapping from the noise code to a tensor that is upsampled and convolved is crucial for disentanglement. By feeding a tensor that is already entangled in terms of the input noise, previous methods struggle when attempting to change individual attributes. We show that it is sufficient to enforce independence between different sources of variation as well as different mapping parameters to obtain disentanglement.\newline 
    We show a setting where foreground-background disentanglement can be improved by borrowing from style transfer methods. Instead of employing labeled alpha masks, we simply encourage one code to define the part of the image that is related to content, and another code to define the part of the image that is related to style.\newline
    In terms of application, we overcome the current challenges for face image manipulation. Our method produces high quality images without requiring labels or complicated architectures. Moreover, we relax the constraints on the degree of control, giving users the freedom to change almost every aspect of generated images.\par
    The main limitation of our work is the inability to change the background without changing anything about the foreground. While we are able to change the foreground while keeping the same background, changing the background (style code) will change color and sometimes features about the original face. Another open area of research is eliminating the remaining entanglement between mid-level features, so that gender can be specifically changed without changing age. There is also some tuning required to achieve the desired granularity of control. Though disentanglement is often achieved regardless of the particular configurations, the exact effects vary depending on the noise injection settings.\par
    Nonetheless, we believe that our work will lead to a wider adoption of GANs for artistic and task-dependent image generation due to the simplicity of the implementation and the quality of the results. Disentangled control over generated images has been a major problem since the introduction of GANs. We have taken large strides towards solving these problems. Our work opens the door for understanding the representation learned by GANs; why they are able to generate different poses, hairstyles, and facial expressions of the same person even though neither the concept of identity nor labeled attributes are present in the dataset.
{\small
\bibliographystyle{ieee_fullname}
\bibliography{egbib}
}

\end{document}